\documentclass[letterpaper, 10pt, conference]{ieeeconf}
\IEEEoverridecommandlockouts                              

\usepackage{xcolor}
\usepackage{tabularx}
\usepackage{amssymb}
\usepackage{array} 
\usepackage{arydshln}
\usepackage{amsmath}
\usepackage[sorting=none]{biblatex}
\usepackage{graphicx}
\usepackage{multirow} 
\usepackage{booktabs} 
\usepackage{dsfont}
\usepackage{threeparttable} 
\usepackage{multicol}

\begin{document}

\title{Revisiting Reward Design and Evaluation for \\ 
Robust Humanoid Standing and Walking}

\author{
Bart van Marum,
Aayam Shrestha,
Helei Duan, 
Pranay Dugar, 
Jeremy Dao, 
Alan Fern
   \thanks{All authors are with the Dynamic Robotics and Artificial Intelligence Laboratory. Oregon State University. Corvallis, Oregon, USA. Email: \texttt{bart@vmeps.com, shrestaa, duanh, dugarp, daoje, afern@oregonstate.edu.}}%
}


\maketitle

\begin{abstract}
A necessary capability for humanoid robots is the ability to stand and walk while rejecting natural disturbances. Recent progress has been made using sim-to-real reinforcement learning (RL) to train such locomotion controllers, with approaches differing mainly in their reward functions. However, prior works lack a clear method to systematically test new reward functions and compare controller performance through repeatable experiments. This limits our understanding of the trade-offs between approaches and hinders progress. To address this, we propose a low-cost, quantitative benchmarking method to evaluate and compare the real-world performance of standing and walking (SaW) controllers on metrics like command following, disturbance recovery, and energy efficiency. We also revisit reward function design and construct a minimally constraining reward function to train SaW controllers. We experimentally verify that our benchmarking framework can identify areas for improvement, which can be systematically addressed to enhance the policies. We also compare our new controller to state-of-the-art controllers on the Digit humanoid robot. The results provide clear quantitative trade-offs among the controllers and suggest directions for future improvements to the reward functions and expansion of the benchmarks.

\end{abstract}
\IEEEpeerreviewmaketitle


\section{Introduction}
\begin{figure}[th!]
    \centering
    \includegraphics[width=0.9\columnwidth]{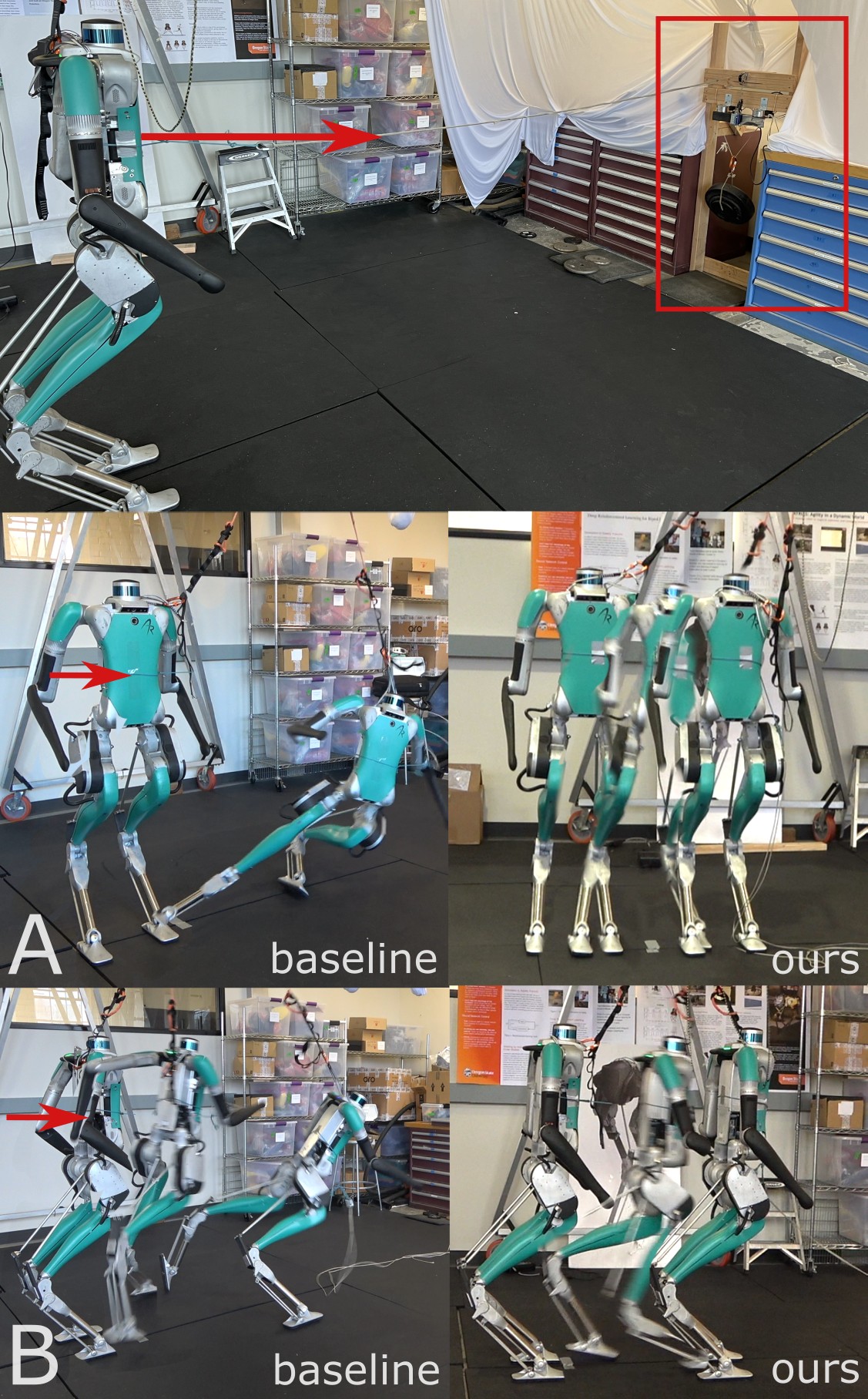}
    \caption{We propose a set of metrics with an easy-to-setup testing fixture and provide quantitative results towards the controller performance in the real-world. Our proposed RL-based method produces a robust standing-and-walking controller for the humanoid robot Digit. The learned controller can handle a set of significant amount of disturbances, such as lateral push at 150N for 500ms shown in A and sagittal push at 200N for 500ms shown in B. The controller is able to walk, stand, and seamlessly transition between these two settings.}
    \label{fig:enter-label}
    \vspace{-1em} 
\end{figure}

Humanoid robots hold the promise to deliver enormous amounts of human-like physical labor across varied real-world spaces. A precondition for realizing this potential is for humanoids to be able to stand and walk (SaW) in natural settings while withstanding typical disturbances encountered during deployments. In particular, walking provides mobility and standing provides the foundations for humanoid manipulation tasks. For more traditional wheeled based robots, designing controllers for  analogous SaW behaviors is relatively straightforward due to the inherent stability of wheeled platforms. However, for humanoids, even these basic SaW behaviors are a challenge due to the inherent instability of bipedal systems.  

There has been recent progress in bipedal locomotion through sim-to-real reinforcement learning (RL)~\cite{ogCassieRL, siekmann2021sim, li2021reinforcement,castillo2023template,dao2022sim,duan2023learning}. The approaches train controllers in simulation using various reward functions to shape the locomotion behavior while performing domain randomization to help transfer to the real world. While these approaches have yielded impressive real-world video demonstrations, there have not been repeatable quantitative evaluations that allow for clear comparison of the real-world trade-offs among different methods. This limits our ability to systematically explore the space of reward functions to develop effective SaW controllers. The lack of systematic evaluation is especially pressing as current reward functions are often based on engineered constraints which may be counter-productive for producing reliable SaW behaviors under disturbances.

In this paper, we address the above issues with the goal paving the way for experimentally measurable progress in SaW control learning. Our first contribution is to introduce a SaW benchmarking procedure that specifies repeatable physical experiments for measuring real-world metrics of SaW controllers. Importantly, these experiments can be conducted without expensive equipment, using simple devices that can be constructed from readily available materials. As our second contribution, we leverage the proposed benchmark to clearly identify areas of improvement for current reward functions, leading us to revisit the reward-design problem for SaW controllers. We attempt to construct a minimally-constraining reward function that avoids undesired behavior with the hope that such a reward function will allow for more flexibility in learning to reject disturbances and react to changing commands. Our third contribution is to use the benchmarking procedure to evaluate and compare our new controller with the manufacturer-supplied controller and a current state-of-the-art reward function. The results reveal trade-offs that would not be apparent without the rigor of repeatable experiments, emphasizing the importance of systematic evaluation in advancing the field of SaW control learning.

Importantly, we are not claiming that the current SaW benchmarks are complete or that the new reward function is truly minimal and cannot be further improved. Instead, we view the SaW benchmarks and new reward function as starting points on a trajectory of continual, measurable real-world improvements. Indeed, our experiments in this paper already point to directions for improvement.


\section{Problem Statement and Related Work}
We consider the problem of producing a controller for a bipedal humanoid robot that supports the following two commands: \textbf{1) Stand.} The robot should stop if moving and stand in place with two feet on the ground. \textbf{2) Walk.} The robot should walk at a specified velocity (direction and speed) and a specified heading with an important special case corresponding to rotating in place. Note that the key distinction between standing and walking is that for standing the default behavior should be for both feet to be on the ground, while for walking, including walking in place, the feet cycle through stance and swing phases. 

To be useful in practice, a \emph{standing and walking (SaW)} controller must be able to reliably switch between different commands and reject physical disturbances, such as bumps or terrain features, that may occur in an application. The primary objective of disturbance rejection is to prevent the robot from falling, while the secondary objective is to minimize departure from the commanded behavior. For example, when commanded to stand the occurrence of a large enough disturbance may require the robot to take a step in order to avoid falling, which conflicts with the default command objective of having two feet on the ground. However, we would typically prefer the robot to take the step, rather than risk falling by struggling to keep both feet planted.

\textbf{Prior Standing and Walking Control.} There have been a number of prior approaches for model-based humanoid SaW control~\cite{kuindersma2016optimization, feng2014optimization, apgar2018fast, gong2019feedback, xiong20223}. More recently, there has been significant progress in using sim-to-real RL for training bipedal locomotion controllers, e.g. blind locomotion for multiple gaits~\cite{siekmann2021sim, li2021reinforcement, castillo2023template}, locomotion under different loads~\cite{dao2022sim}, and visually-guided locomotion over irregular terrain~\cite{duan2023learning}. In many cases, these controllers do not have a native standing mode and instead walk in place. In cases where standing is explicitly supported, a separate standing controller is usually designed or trained (e.g.~\cite{crowley2023optimizing, dao2023sim, castillo2021robust}). 

One challenge in using separate standing and walking controllers is that switching between the control modes is not always straightforward. This is because the control state-spaces of each controller have little native overlap, which requires the use of difficult to tune heuristics to yield smoother and more reliable transitions. For example, this may involve hand-engineered cycle-time constraints in transitioning from waking to standing~\cite{crowley2023optimizing} or blending the outputs of the controllers at transition points~\cite{dao2023sim, tidd2022learning}. Another approach to improving controller switching is to train controllers for different behaviors from starting states typical of other behaviors~\cite{dao2023sim}, which raises the challenge of generating appropriate start state distributions. Alternatively, in this work, we choose to train a single SaW controller that does not require any form of reference or clock-based inputs, and natively learns to switch freely between commands throughout the entire learning process.

\textbf{Reward Functions for Bipedal Locomotion.} One of the challenges in sim-to-real RL for bipedal locomotion is specifying a reward function that results in the desired locomotion characteristics. Prior work has used highly prescriptive reward functions to describe preferred characteristics of the behavior to be learned. This includes clock-based rewards that asserts preferences for stance and swing phases in alignment with periodic clocks~\cite{siekmann2021sim}, and reference-motion rewards that attempt to imitate joint or task space reference trajectories captured from humans~\cite{tang2023humanmimic, cheng2024expressive} or produced through optimization~\cite{green2021learning}.  Designing such prescriptive reward functions is expensive and brittle as they must be retargeted to each robot and behavior variation (e.g. velocity) or extensively tuned and do not handle transitions between gaits well. For example, prior work on fast locomotion (i.e. running) requiring substantial manual tuning for transitions between multiple speed gaits and standing~\cite{crowley2023optimizing}. While such prescriptive reward designs have yielded impressive controllers it is unclear whether they are placing unnecessary, or even counter-productive constraints, on the learned controllers. For example, handling disturbances can require generating behavior that significantly departs from the constraints imposed by the periodic clock or reference motions. These challenges provide our motivation for investigating minimally-constraining reward functions in this paper.

\textbf{Evaluating Locomotion Controllers.} The vast majority of prior work on model-based and RL-based control for bipedal locomotion does not report real-world performance metrics based on repeatable experiments. Instead, the primary method of demonstrating real-world performance is based on producing videos of the controllers in action. For example, many studies are limited to apply external disturbances by human experimenters~\cite{li2021reinforcement, radosavovic2023learning} or demonstrating the robot traversing arbitrary terrains and obstacles~\cite{siekmann2021sim}, which are non-reproducible and not quantitative. Some exceptions do consider hardware setups for repeatable tests of specific tasks~\cite{9981359}, such as disturbance rejection and low-level velocity tracking. For example, prior work~\cite{weng2023standardized} used an impact generator, similar to those used in vehicle testing, which provides repeatable quantitative results, but is not widely available to researchers. The failure to establish objective measurements of real-world performance that can be easily implemented by most researchers has made it difficult to judge progress and compare competing methodologies. This failure motivates our proposal in this paper for simple metrics that can be experimentally measured using widely available materials and devices.


\section{Quantitative SaW Performance Benchmark}
We propose a reproducible set of benchmarks for quantitatively assessing key aspects of a SaW controller in the real-world. These metrics quantify the disturbance rejection ability, accuracy in command following, and energy efficiency. The benchmark is intended to allow comparison of any SaW controller, regardless of what method a controller is based on.

\subsection{Disturbance Rejection}
To enable standardized comparison, we propose a formal impulse perturbation test with precise control over disturbance parameters. Specifically, our method applies a force $\boldsymbol{F}$ of fixed magnitude and duration $\Delta t$ to the robot base, generating an impulse $\boldsymbol{J}$. By systematically varying $\boldsymbol{F}$ and $\Delta t$ over a range of values, we can quantify performance in terms of the success rate for recovery after experiencing the impulse $\boldsymbol{J}$.

In order to objectively measure this metric in the real world it is necessary to apply forces consistently in a repeatable manner, without human input. To that end we construct a simple impulse application device from off the shelf materials.

\begin{figure}
    \centering
    \includegraphics[width=0.98\columnwidth]{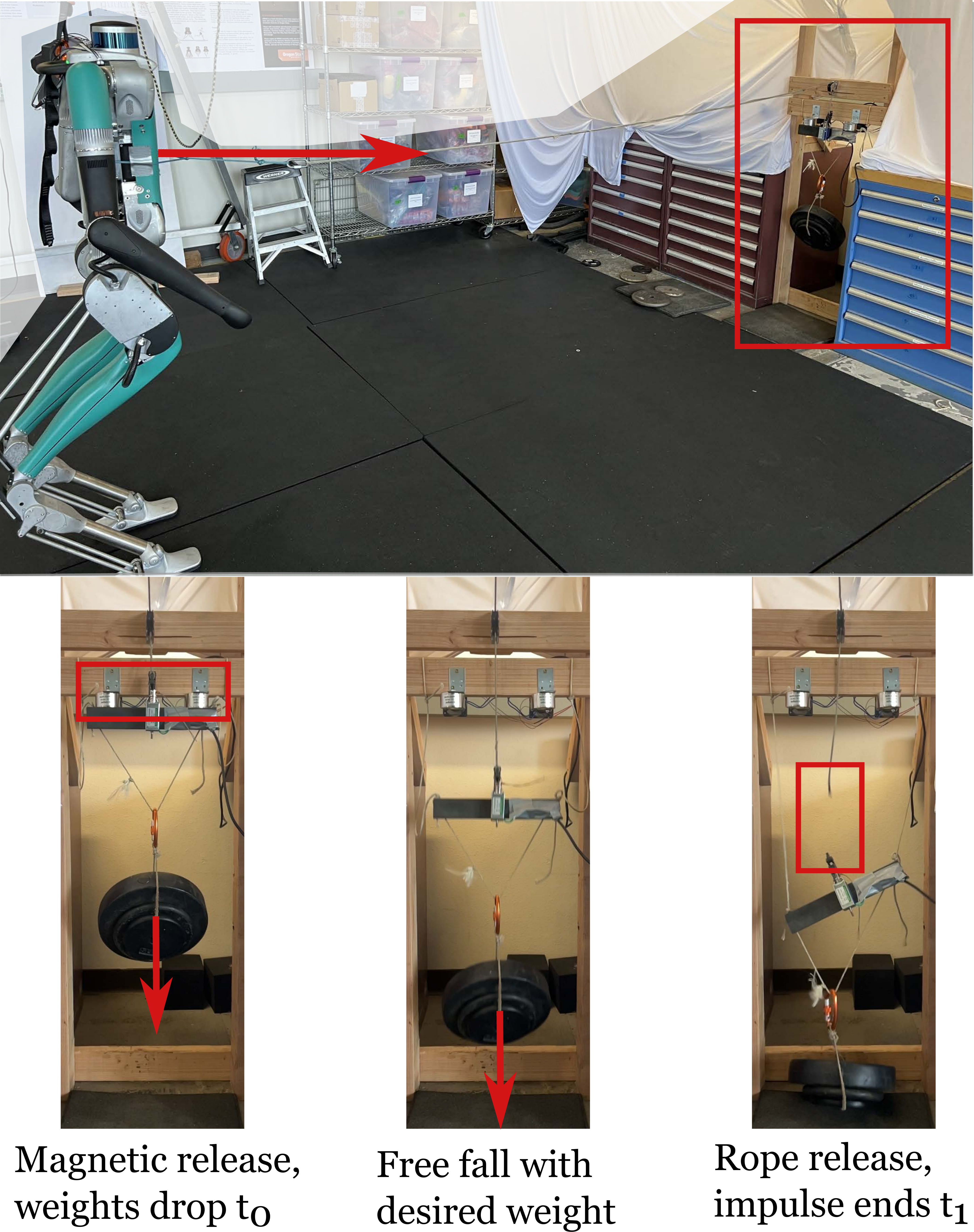}
    \caption{An impulse is applied to the robot by means of a weight connected by a rope. Force $\boldsymbol{F}$ is regulated by adding and removing weight. Duration $\Delta t$ is regulated by a microcontroller that automatically disconnects the weight from the rope, after a set amount of time. The rope is always attached to Digit at the same height of 122 cm.}
    \label{fig:force_applicator}
    \vspace{-1em}
\end{figure}

Figure~\ref{fig:force_applicator} shows our impulse application device in the lab environment. The impulse applicator works by releasing a weight suspended by magnets, which is automatically disconnected after a preset duration. After applying a fixed duration impulse, the robot is freely able to recover. We provide a video\footnote{https://youtu.be/ZraDSE5Peeg?si=Pt9xwpN2SNihLNTp\&t=46} to further clarify this procedure and have open-sourced the part list and construction details on the project website\footnote{https://b-vm.github.io/Robust-SaW/}.

\textbf{Metric 1: Standing Fall Percentage.} In this paper, to simplify the measurement setup, we limit the experimental scope to characterizing disturbance rejection when standing. We focus on disturbances in the positive and negative X and Y directions, where the positive X direction is the direction faced by the robot. For each direction and selected combinations of weight and duration, we compute the metric value over multiple trials. Each trial involves initializing the robot by issuing a standing command and then using the device connected in the appropriate direction to provide the specified impulse weight and duration. The metric value is the percentage of trials leading to success, where a trial is successful if the robot does not fall. There are many variations of this metric using the same device that can be considered in future work. For example, defining more detailed metrics such as how much the average robot position changed after the disturbance, the time taken for recovery, or the same procedure can be applied to a robot when walking at a particular speed.  \\

\subsection{Command Following}
Precisely executing commanded motions is essential for a humanoid robot to reliably perform planned behaviors. However, metrics of command following accuracy often rely on motion capture systems, limiting reproducibility.
To address this, we propose three simple metrics that measure the accuracy of velocity control and control of in-place robot rotation. 

\textbf{Metrics 2 and 3: In-Place Rotation Accuracy.} For certain applications it is useful for a humanoid to be able to rotate its body to a particular orientation while remaining in place. To test this we conduct trials where the robot starts in a standing position in the middle of a 2ft diameter circle, which is considered the region of zero positional error. The robot is then given a command to rotate at angular velocity $\omega_z$ for $\Delta t$ seconds, which ideally should correspond to a commanded orientation change of  $\theta_c = \omega_z \cdot \Delta t$. We compute two metrics at the end of the trial: 1) \emph{angular error}, which is the difference between the commanded angular rotation and the actual rotation, and 2) \emph{lateral drift}, which is measured as the distance of the furthest foot from the boundary of the starting circle. 

\textbf{Metric 4: Velocity Accuracy.} We consider a simple velocity tracking test using only basic measurements. Specifically, we issue a constant velocity command $v$ for duration $\Delta t$, which should ideally produce a net translation of $d_c = v \cdot \Delta t$. For our current procedure, each trial of the experiments starts the duration clock when the robot is in a standing position and then after $\Delta t$ seconds the standing command is issued. The distance traveled is then manually measured and averaged across trials. By comparing the actual distance traveled $d_r$ to the commanded distance $d_c$, we can quantify velocity control performance without specialized equipment. This metric assumes that the SaW controller under test is able to walk on flat terrain without falling. Note that, this procedure does not explicitly account for startup and stopping times for the robot. Instead, velocity ramp up and down are assumed symmetrical, thus canceling out.

\subsection{Energy Efficiency}
\textbf{Metric 5: Energy Efficiency.} In addition to quantifying the accuracy of command execution, energy efficiency of command execution is an important performance metric for real-world humanoid viability. More efficient gaits directly extend operational runtime by conserving battery power. Additionally, a more efficient gait reduces mechanical wear from torque and impacts, prolonging hardware lifespan. The energy usage of a gait over a duration $\Delta$ can be approximated by computing the work done by the actuators: $W = \int_0^{\Delta} \boldsymbol{\tau}_t \cdot \boldsymbol{\omega}_t \, dt$, where $\boldsymbol{\tau}_t$ represents the motor torques at time $t$ with the motor angular velocities $\boldsymbol{\omega}_t$. This metric ignores motor losses, particularly significant when standing, so we omit standing measurements. In our experiments we estimate the work done over a time interval by using the robot's estimated torques, rather than the actual torques, due to the lack of a torque sensor and current draw information. As a metric we report the estimated average energy usage per meter traveled to normalize for different travel distances.


\section{SaW Training and Reward Design}

Below, we first describe our controller architecture and simulation-based training framework. Next, we describe the design of our SaW reward function that aims to minimally constrain the learned behavior. 

\subsection{Architecture and Training Framework}

Our SaW controller is a (64, 64) two-layer long short-term memory (LSTM) ~\cite{lstm} recurrent neural network. The input to the network consists of: 1) the current robot state [\textit{motor velocities}, \textit{motor positions}, \textit{joint velocities}, \textit{joint positions}, \textit{torso orientation}], and 2) the user command~$c_u =[c_x, c_y, c_{yaw}]$, specifying the desired x, y, and angular velocities, where standing corresponds to $c_u = [0, 0, 0]$. The output action of the controller is the joint-space PD setpoints for all 20 actuators. The SaW controller is operated at 50Hz with the PD controllers operating at 2kHz. We train the controller using the Proximal Policy Optimization (PPO)~\cite{schulman2017ppo} RL algorithm, which is extended with a mirror loss, to encourage symmetry in policy behavior. Our domain randomization is similar to \cite{duan2023learning}.

We follow a standard episodic RL training process. Each episode begins with the robot initialized in a default standing position and lasts for 16 seconds of simulated time, or is terminated early when the robot falls. During an episode a new user command is uniformly sampled every $2$ to $6$ seconds from five categories: [\textit{standing}, \textit{walking in sagittal plane}, \textit{walking laterally}, \textit{rotating in place}, \textit{omnidirectional walking}]. User command ranges are sampled from $c_x$:~$[-0.5, 2.0]$~m/s, $c_y$:~$[-0.5, 0.5]$~m/s, $c_{yaw}$:~$[-0.5, 0.5]$~rad/s.

In addition, during each episode, for each frame there is a 1\% chance of getting a random push to encourage disturbance rejection capability. These random pushes were uniformly sampled from a range of 200N to 800N, lasting a single timestep (20ms). Our early ad-hoc testing indicated that using a more diverse distribution did not improve performance. However, as we will see in our experiments, our real-world benchmarking process will lead to a different conclusion (see Section \ref{sec:improvement}).

\subsection{Reward Design}

During an episode, we seek a reward function that indicates throughout an episode how well the behavior satisfies the current user command. Additionally, the reward function should achieve this, while also minimally constraining other aspects of the learned behavior. Our design process began with an enumeration of the reward terms found in the many prior studies on RL for legged-robot locomotion. We then progressively added terms based on observations of intermediate learned policies. The resulting reward function is composed of a number of additively weighted reward terms, which are shown in Table ~\ref{table:reward} along with the associated weights. Below we outline the essential role that each term plays in the overall design. 

\textbf{Basic Command Following.} The first three essential components in Table ~\ref{table:reward} measure how well the current robot velocities and orientation match the commands. We found that training with just these components results in a hopping locomotion behavior, where the robot moves by jumping with both feet. While this behavior satisfies the commands, it is not desirable walking behavior indicating that additional reward terms are required.  

\textbf{Single Foot Contact.} To address hopping we have found multiple approaches that can individually be added to the above three reward terms to learn to walk instead. These include: 1) adding a base height reward, 2) clock based rewards and inputs~\cite{allgaitsclocks}, 3) tuning exploration noise, 4) a feet contact transition reward, and 5) a single feet contact reward. We found that the most reliable and unconstrained way to produce walking instead of hopping is via the single foot contact reward, which also does not require tuning. 

For non-standing commands, the single foot contact component provides a reward of 1 at each time step where only one foot is in contact with the ground. To allow for some overlap in the stance and swing phases, we add a grace period of $0.2$ seconds. This means that if single contact occurred at least once in the last $0.2$ seconds, the reward is granted, otherwise the reward is 0. 

For the standing command, this reward component is a constant of 1, giving no preference for foot contact. Intuitively, we might expect standing to involve rewarding double foot contact. However, this is problematic since it penalizes the recovery steps needed to reject disturbances, as that requires breaking ground contact of at least one of the feet. Additionally, when transitioning from walking to standing, requiring double foot contact will cause a policy to opt for the closest stance position rather than the most stable one. Thus, to learn standing while avoiding these problems we opt to implicitly reward standing utilizing existing reward terms. It turns out that most reward terms will be greater when a policy stands still with both feet on the ground, than when it steps in place or only stands on one foot. 

\textbf{Sim-to-Real and Style.} While the combination of the above reward terms are enough to reliably train a SaW controller in simulation, RL often results in controllers that lack certain stylistic characteristics that can influence visual properties and more importantly stability of sim-to-real transfer. We note that, it is not clear whether these stylistic characteristics are missing due to the inability of RL to perfectly optimize the above reward terms, or if the above terms are inherently inadequate. In either case, the inclusion of additional stylistic reward terms is intended to more reliably and quickly lead RL to acceptable solutions. The following terms are ones that we have found to be important to include, noting that a more exhaustive analysis may reveal a more minimal set.
\begin{itemize}
    \item \emph{base height}: maintain a consistent base height during different modes and commands. 
    \item \emph{feet air time}: 
    this term regularizes the stepping frequency by applying a penalty of 0.4 at each foot touchdown, which can be counteracted by a positive reward component equal to the number of seconds since the foot has been in the air (airtime). Without this component the learned controllers tend to favor gaits with step frequencies that are stylistically too large, which may be due to those frequencies corresponding to likely local minima. This component is a constant when standing. 
    \item \emph{feet orientation}: make feet point out straight at all times, except when commanded to rotate.
    \item \emph{feet position}: loosely define the foot position during stance, to prevent undesired stance positions.
    \item \emph{arm position}: loosely defines desired arm joint angles to prevent unwanted arm movements, while allowing use of arms for balancing. This stylistic component can be important to minimize the potential for self-collisions. 
    \item \emph{base acceleration}: prevent jerky base motions. 
    \\

    \item \emph{action difference}: minimize changes in actions between timesteps. 
    \item \emph{torque}: minimize torque usage. 
\end{itemize}

\vspace{1em}

\textbf{No Clocks.} While prior work that uses clock-based reward signals (e.g. ~\cite{allgaitsclocks}) does allow for standing, the nontrivial question of what to do with the required clock inputs during standing and transitions remains a challenge. Additionally, the clock framework incentivizes low foot velocities in standing mode, which directly impedes disturbance rejection capabilities. Rather, the above reward function does not require reference clocks, trajectories or signals of any sort to learn walking, and allows a policy to control such parameters internally. Without such signals we eliminate the problem of having to engineer the transitions between standing and walking modes. Additionally for disturbance rejection, without any references to attain to, the policy is free to move feet in any way it seems fit to stay upright.

\begin{table}[t]
  \centering
  \begin{threeparttable}
    \caption{Reward Terms}
\centering
\scriptsize
\begin{tabular*}{0.48\textwidth}{llc}
\hline
\textbf{Reward Term} & \textbf{Definition} & \textbf{Weighting}  \\
\hline

\addlinespace[2pt]
$x, y$ velocity &
\(\begin{cases} 
e^{-5\cdot(v_{xy} - c_{xy})} & \text{if } c_s \\
e^{-5\cdot(v_{xy} - c_{xy})^2} & \text{else }
\end{cases}\)
& 0.15, 0.15 \\

Yaw orient. & \(e^{-300\cdot qd(\mathbf{q}_{yaw}, \mathbf{c}_{yaw})}\) & 0.1 \\

Roll, pitch orient. & \(e^{-30\cdot qd(\mathbf{q}_{rp}, \mathbf{c}_{rp})}\) & 0.2 \\

Feet contact & 
\(\begin{cases} 
    1 & \text{if } c_s \\
    1 & \text{if } n_{c, t^{*}} = 1  \text{ for any } t^{*} \in [t-0.2, t]\\
    0 & \text{else} 
\end{cases}\)
& 0.1 \\
\addlinespace[2pt]

Base height & \(e^{-20\cdot|p_z - c_{h}|}\) & 0.05 \\

Feet airtime &
\(\begin{cases} 
    1 & \text{if } c_s \\
    \sum_{f\in (l,r)}(t_{air, f}-0.4) * \mathds{1}_{td, f} & \text{else}
\end{cases}\)
& $1.0^\dag$ \\
\addlinespace[2pt]

Feet orientation &
\(\begin{cases} 
e^{-\sum |\mathbf{r}_{\text{feet, rp}} - \mathbf{c}_{feet, \text{rp}}|} & \text{if } |c_{yaw}| > 0 \\
e^{-\sum |\mathbf{r}_{\text{feet, rpy}} - \mathbf{c}_{feet, \text{rpy}}|} & \text{else}
\end{cases}\)
& 0.05 \\

Feet position & 
\(\begin{cases} 
e^{-3 \cdot |\mathbf{p}_{feet} - \mathbf{c}_{feet}|} & \text{if } c_s \\
1 & \text{else}
\end{cases}\)
& 0.05 \\
\addlinespace[2pt]

Arm & $e^{-3 \cdot ||\boldsymbol{\theta}_{\text{arm}} - \boldsymbol{c}_{\text{arm}}||}$ & 0.03 \\

Base acceleration & $e^{-0.01 \cdot \sum|\mathbf{b}_{xyz}|}$ & 0.1 \\
Action difference & $e^{-0.02 \cdot \sum |\mathbf{a}_{t} - \mathbf{a}_{t-1}|}$ & 0.02 \\
Torque & \(e^{-0.02 \cdot \frac{1}{N} \sum |\mathbf{t}_{\text{motor}}|/\mathbf{t}_{\text{max}}}\) & 0.02 \\

\hline
\end{tabular*}
\label{table:reward}
    \begin{tablenotes}\footnotesize
    \item $c$ = a command;
    $c_s$ = standing command; 
    $q$ = a quaternion; 
    $p$ = a position;
    $\boldsymbol{b}$ = base acceleration;
    $qd(\cdot)$ = quaternion distance function;
    $n_{c}$ = number of feet in contact with ground;
    $\mathds{1}_{td}$ = boolean variable indicating a touchdown in the current timestep;
    $^\dag$ =  note that the feet airtime reward is the only sparse reward, therefore the weight is significantly higher than other terms. 
    \end{tablenotes}
  \end{threeparttable}
  \vspace{-1em} 
\end{table}


\section{Evaluation Results} 
\begin{figure*}[ht]
    \centering
    \includegraphics[width=\columnwidth]{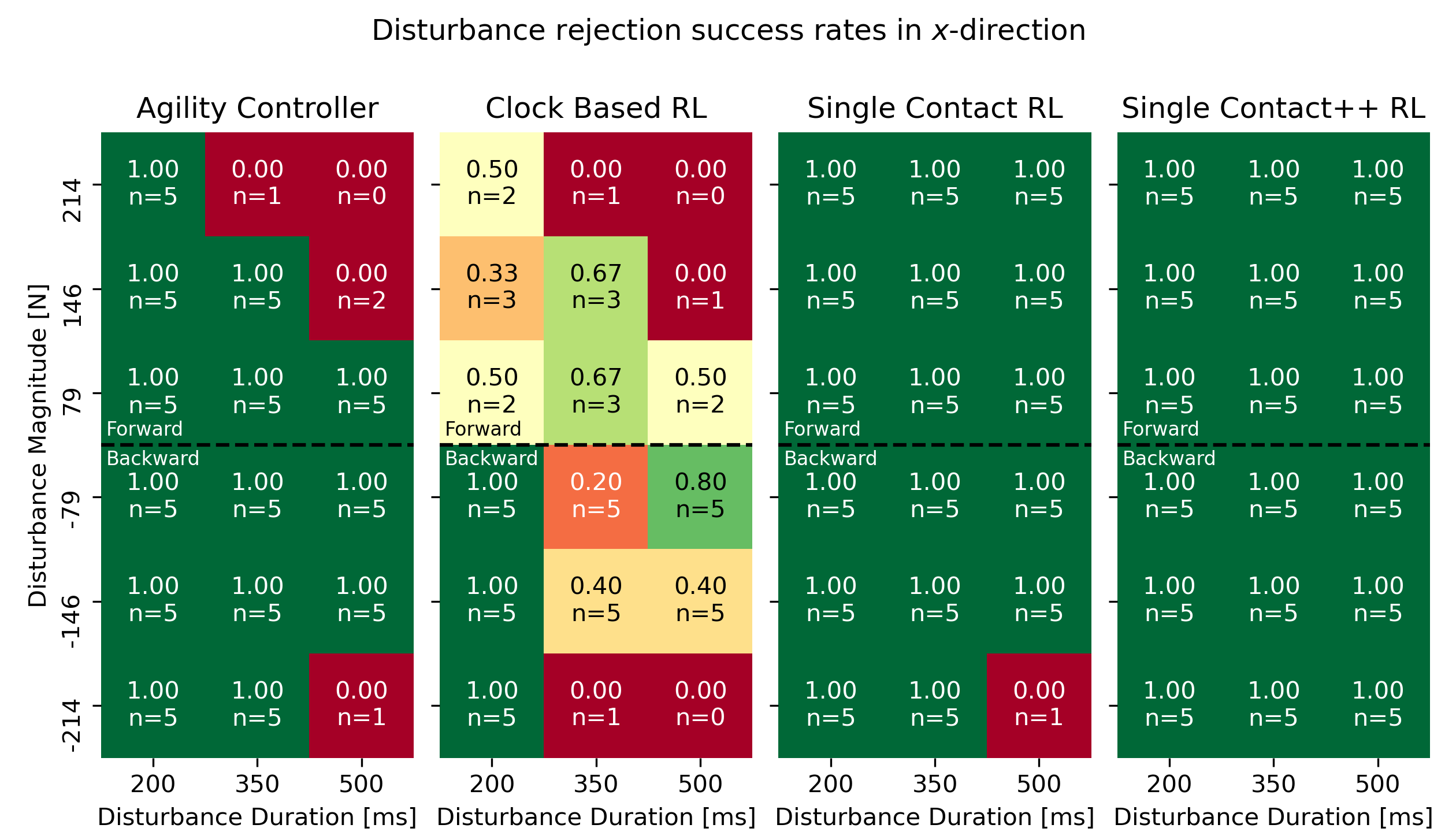}  \hspace{0.75em} \includegraphics[width=\columnwidth]{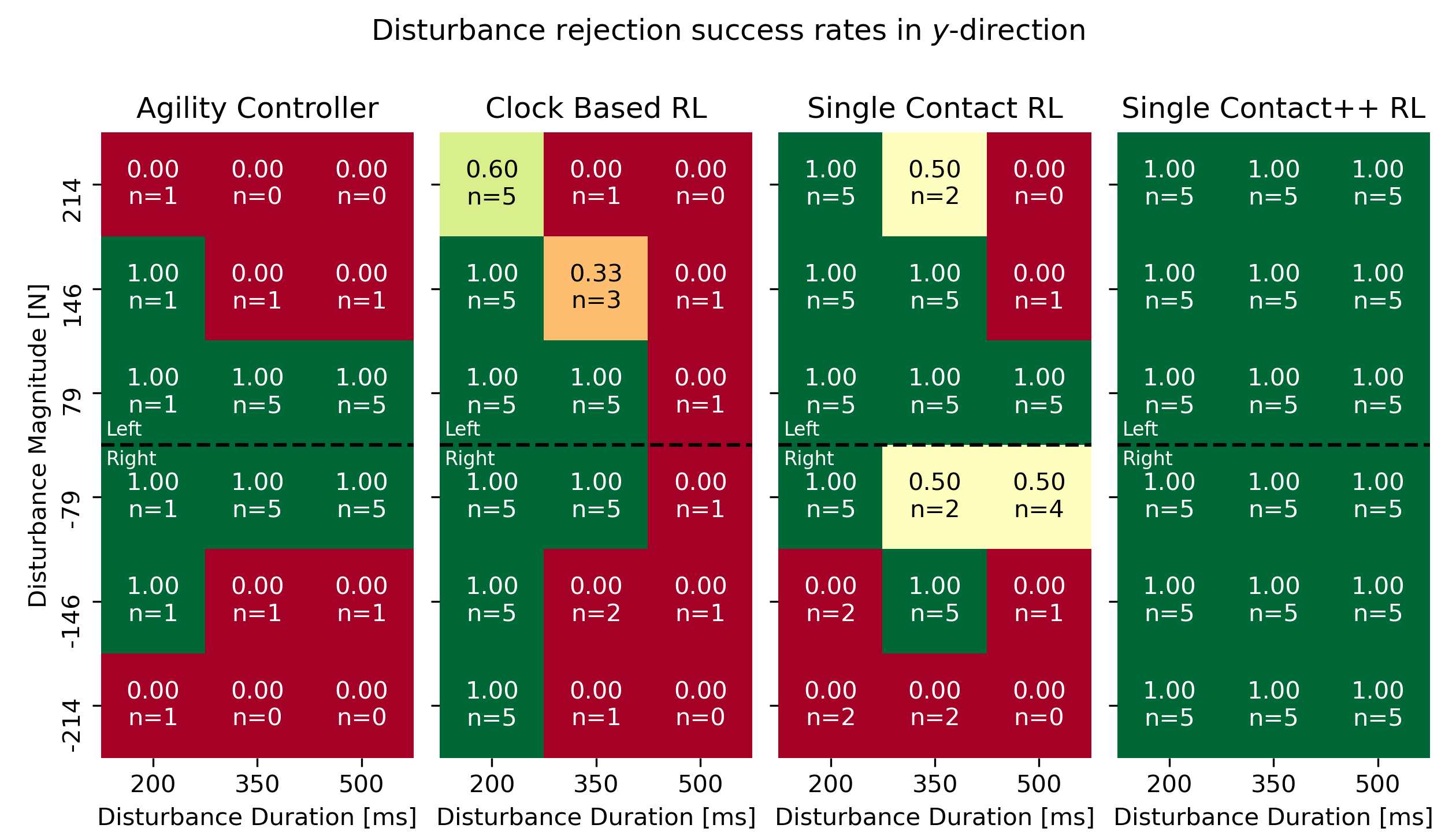}
    \caption{Disturbance rejection success rates for various humanoid SaW controllers in the $x$-direction (left) and $y$-direction (right). Results show that our Single Contact++ reward function outperforms competing alternatives. Our Single Contact controller shows asymmetric and non-monotonic results in the $y$-direction, emphasizing the importance of systematic evaluation.}
    \label{fig:distr_xy}
    \vspace{-1em} 
\end{figure*}

In this section, we use our proposed benchmarking procedure to evaluate and compare three SaW controllers for the Digit V3 humanoid robot manufactured by Agility Robots: 1) \emph{Single Contact RL.} trained using our minimally-constrained SaW reward function from Table \ref{table:reward}, 2) \emph{Clock Based RL.} trained using a state-of-the-art clock-based~\cite{allgaitsclocks} reward function, and 3) \emph{Agility Controller.} the manufacturer-provided controller. In addition, we describe how the results motivated the training of a new SaW controller, called \emph{Single Contact++ RL}, which yielded measurable improvement. 

\subsection{Disturbance Rejection While Standing}

\textbf{Metric 1: Standing Fall Percentage.} We use our impulse application setup to conduct disturbance rejection experiments on the robot. For each impulse, we attempt 5 trials to gauge reliability and stop testing at the first failure to avoid damage to the robot. Figure~\ref{fig:distr_xy} show results for $x$ and $y$ directions respectively. We note the number of tests for every impulse in the plots. The range for force and duration is $[79, 214]$~N and $[200, 500]$~ms. We chose these ranges of forces and durations to yield a set of impulses that are both challenging to all controllers, while minimizing risk of damage to the hardware. The upper end of the force range corresponds to approximately half the robot weight. 

We find that the Single Contact RL Controller outperforms the Clock based controller in the x-direction, for both forward and backward pushes, as expected. Interestingly there are large asymmetries and gaps showing up for the Single Contact RL controller in the y-direction. Unintuitively, in some cases a higher impulse push can be more easily dealt with than a lower impulse. This is unexpected, and underscores that claims of stability must be substantiated through formal experiments.

\subsection{Benchmark Guided Improvement}
\label{sec:improvement}

The above results revealed several weaknesses in the disturbance rejection profile of the Single Contact RL controller. In particular, the maximum backward force and duration led to complete failure and there were many gaps in the y-direction disturbance profile. These results led us to hypothesize that the limited disturbances used in our simulation training episodes were not adequate for the desired sim-to-real robustness. We trained a new version of Single Contact RL, called \emph{Single Contact++ RL}, by using the same reward function but adjusting the simulated disturbances to be closer to the test distribution. Specifically, the random pushes are uniformly sampled from [20N, 200N] and [200ms, 500ms] uniformly in all three dimensions. In addition, in an attempt to encourage improved drift performance we increased the duration of commands before a new commands is sampled from [40, 100] timesteps to [100, 300] timesteps. 

The results of disturbance benchmarking for this new controller are in Figure~\ref{fig:distr_xy} and show significant improvement, achieving perfect disturbance rejection in our tests. In an attempt to find the limits of the \emph{Single Contact++ RL} controller we performed additional disturbance tests of 258~N @ 500~ms, the max our pulling device is capable of. The \emph{Single Contact++ RL} controller passed this test in all directions. We do not investigate this further due to the risk of damaging the hardware. This result shows the utility of the proposed benchmarking approach for driving measurable improvement.

\subsection{Command Following}
\begin{figure}[t]
    \centering
    \includegraphics[width=\columnwidth]{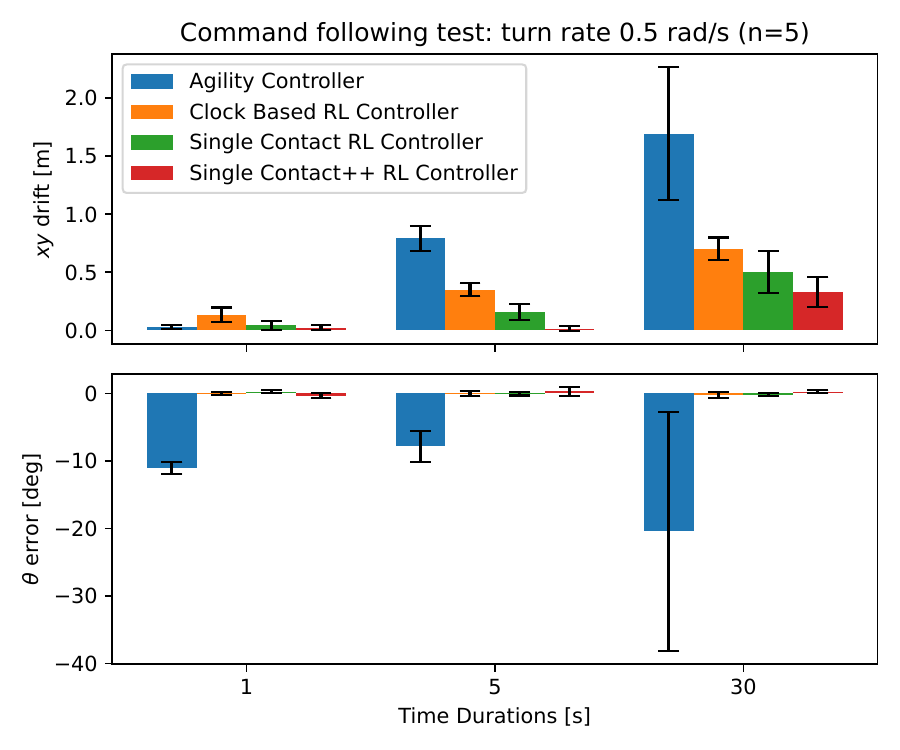}
    \caption{Command following accuracy for turning in place. Error bars are standard deviation. Also note that the 30 seconds drift results for Agility Controller were in some cases helped by the robot tether. It is safe to assume results without tether would have been closer to the upper end of the error bar.
    }
    \label{fig:rz_results}
    \vspace{-1em}
\end{figure}

\textbf{Metrics 2 and 3: In-Place Rotation Accuracy.} We measured the drift and heading accuracy for all controllers rotating at 0.5~rad/s for 1, 5 and 30 seconds. The results in Figure \ref{fig:rz_results} reveal substantially lower drift for the RL controllers compared to the model-based Agility Controller. The Agility Controller especially suffers large drifts up to 2 meters in the 30 second test. Interestingly, the clock based controller has the largest drift for the 1 second test. We observe from the foot motions that this is caused by sudden changes in the clock signals, due to the transition from standing to walking mode. We observe in our testing that the Single Contact controllers sway less during rotating, causing a lower drift. Finally, the Single Contact++ controller performs best in all cases. We hypothesize this is due to the longer time range used for command windows during its training, which allows the policy to become more sensitive to drift that accumulates over longer time periods.

Figure~\ref{fig:rz_results} also shows the RL controllers have a rather low angular error compared to the Agility controller. This is likely due to an architectural difference, as our controller tracks a rotating heading, rather than a yaw velocity. The result is that the RL controllers are highly accurate in rotating a desired number of degrees. 

\textbf{Metric 4: Velocity Accuracy.} We test velocity command following accuracy by commanding 1~m/s for 10 seconds. The resulting distances are shown in Figure~\ref{fig:eeff}. We find that the Agility Controller undershoots the target, yielding an average velocity of 0.74~m/s. The Clock Based RL controller is most accurate at 1.04~m/s, while the Single Contact RL Controller overshoots at 1.13~m/s. The overshooting performance of the Single Contact RL controller does not occur in simulation, where the actual velocity is slightly below the commanded 1~m/s. This indicates a sim-to-real inconsistency. Interestingly, the increase in speed between sim and real is three times less for the Clock-Based policy, indicating that the constraint imposed by the clock might help this aspect of sim-to-real transfer. This point deserves further investigation. These metrics were not recorded for Single Contact++ due to a hardware failure during experiments close to the submission deadline. 

\subsection{Energy Efficiency}
\begin{figure}[t]
    \centering
    \includegraphics[width=\linewidth]{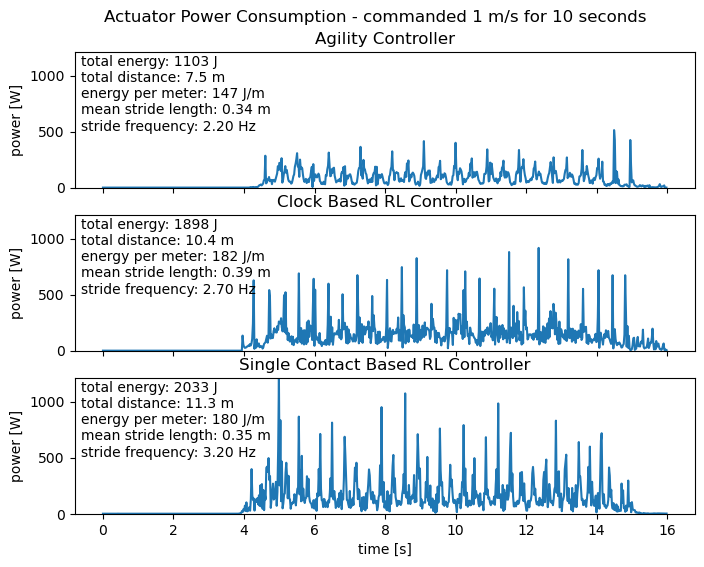}
    \caption{Approximation of power consumption for a commanded run of 1 m/s for 10 seconds. Policies start in standing mode, and end in standing mode. \emph{* Note that results for Single Contact++ RL Controller are missing due to an experiment damaging the robot close to submission.} }
    \label{fig:eeff}
    \vspace{-1em}
\end{figure}

\textbf{Metric 5: Energy Efficiency.} During the same 10 second walk we record positive work done by the motors and plot the power usage estimate in Figure~\ref{fig:eeff}. Interestingly the Agility Controller uses less energy than both RL controllers, when accounting for the difference in distance traveled. This is also apparent visually during tests, as the RL controllers stomp more loudly. Although our method of approximating energy usage is not perfect, it is interesting to realize that at least 33~J/m are not being spent usefully by the RL controllers. We hypothesize that higher energy usage might correspond to the situation where our RL controllers would always anticipate disturbances, so we noticed hardware execution has more stompy touchdowns and more impact forces to be prepared for a disturbance at any time. These results suggest the next iteration of our SaW reward function, which will emphasize smaller ground contact forces. The energy profile was not recorded for Single Contact++ due to a hardware failure close to the submission deadline.  


\section{Summary}
The goal of this paper is to lay the foundations for continually measurable improvement of humanoid standing and walking (SaW) controllers. For this purpose, we introduce a set of quantitative real-world benchmarks for evaluating humanoid standing and walking (SaW) controllers, covering disturbance rejection, command following accuracy, and energy efficiency. We also revisited prior reward designs from previous RL-based SaW approaches in an attempt to arrive at a minimal set of reward terms. We demonstrate the benchmarks' utility by comparing our newly learned SaW controller against a manufacturer-provided controller and a state-of-the-art learning-based controller trained with clock-based rewards. The proposed metrics provided clear, quantitative comparisons of the strengths and weaknesses of these controllers. In particular, the benchmarks reveal unexpected failure modes in the learning-based controller, which guided targeted improvements, ultimately resulting in an enhanced controller that successfully handles all tested disturbances. The benchmark results provide valuable insights into the current limitations of learned controllers for humanoid SaW, including lower energy efficiency and a significant sim-to-real gap.

These results underscore the crucial role of systematic real-world benchmarking in advancing humanoid SaW control and other aspects of humanoid robotic control. The proposed benchmarks and reward functions serve as only a starting point and we believe that by continuously iterating and building upon these benchmarks, the research community can achieve measurable and consistent progress in real-world humanoid locomotion capabilities.

Future work should focus on optimizing energy efficiency without sacrificing performance on other metrics, as well as improving motion smoothness, factors we believe to be correlated.


\section*{Acknowledgment}
This material is based on work supported by the National Science Foundation under Grants 2321851 and IIS-1724360.


\printbibliography


\end{document}